\documentclass[conference]{IEEEtran}
\IEEEoverridecommandlockouts
\usepackage{cite}
\usepackage{amsmath,amssymb,amsfonts}
\usepackage[colorinlistoftodos]{todonotes}

\usepackage{algorithmic}
\usepackage{graphicx}
\usepackage{textcomp}
\usepackage{hyperref}
\hypersetup{colorlinks=true, allcolors=blue}
\usepackage{booktabs}
\usepackage[capitalise]{cleveref}
\usepackage{lineno}
\usepackage{colortbl}
\usepackage{xcolor}
\def\BibTeX{{\rm B\kern-.05em{\sc i\kern-.025em b}\kern-.08em
    T\kern-.1667em\lower.7ex\hbox{E}\kern-.125emX}}
\begin{document}
\def\x{{\mathbf x}}
\def\L{{\cal L}}
\title{TC-GS: Tri-plane based Compression for 3D Gaussian Splatting
\thanks{This work was supported by Guangdong Basic and Applied Basic Research Foundation (Grant No. 2023A1515140037), Guangdong Basic and Applied Basic Research Foundation (Grant No. 2023B0303000010), and Guangdong Research Team for Communication and Sensing Integrated with Intelligent Computing (Project No. 2024KCXTD047).
The computational resources are supported by SongShan Lake HPC Center (SSL-HPC) in Great Bay University.

Corresponding authors: Zitong Yu and Yong Xu.}}
%


\author{
\IEEEauthorblockN{Taorui Wang\textsuperscript{1,2}, Zitong Yu\textsuperscript{2,3}, Yong Xu\textsuperscript{1}}
\IEEEauthorblockA{
\textsuperscript{1}Shenzhen Key Laboratory of Visual Object Detection and Recognition,
Harbin Institute of Technology Shenzhen\\
\textsuperscript{2}School of Computing and Information Technology, Great Bay University\\ 
\textsuperscript{3}Dongguan Key Laboratory for Intelligence and Information Technology\\
}
}

\maketitle

\vspace{-2.5em}

\begin{abstract}
Recently, 3D Gaussian Splatting (3DGS) has emerged as a prominent framework for novel view synthesis, providing high fidelity and rapid rendering speed. However, the substantial data volume of 3DGS and its attributes impede its practical utility, requiring compression techniques for reducing memory cost. Nevertheless, the unorganized shape of 3DGS leads to difficulties in compression.  To formulate unstructured attributes into normative distribution, we propose a well-structured tri-plane to encode Gaussian attributes, leveraging the distribution of attributes for compression. To exploit the correlations among adjacent Gaussians, K-Nearest Neighbors (KNN) is used when decoding Gaussian distribution from the Tri-plane. We also introduce Gaussian position information as a prior of the position-sensitive decoder. Additionally, we incorporate an adaptive wavelet loss, aiming to focus on the high-frequency details as iterations increase. Our approach has achieved results that are comparable to or surpass that of SOTA 3D Gaussians Splatting compression work in extensive experiments across multiple datasets. The codes is available at \href{https://github.com/timwang2001/TC-GS}{TC-GS}. 
\end{abstract}

\section{Introduction}
\label{sec:intro}

Nowadays, novel view synthesis in 3D scene representations has been in a new era thanks to NeRF~\cite{mildenhall2020nerfrepresentingscenesneural}. NeRF and its variants propose rendering colors by accumulating RGB and using a multilayer perceptron (MLP) to predict the attributes of quired points in the 3D scene. While the quality and fidelity are achieved, the expensive querying and MLP slow the rendering progress. To solve this, many approaches have made an effort to enhance training and rendering speed, such as hash grids~\cite{muller2022instant} and parametrization~\cite{barron2021mip,barron2022mip,hu2023tri}. However, they still face relatively slow rendering speeds due to frequent ray point sampling.

Recently, 3D Gaussian splatting (3DGS)~\cite{kerbl20233d} has been proposed as an efficient technique for 3D scene representation and achieved state-of-the-art (SOTA) rendering quality and speed. As an emerging alternative strategy for representing 3D scenes, 3DGS represents a 3D scene using a set of neural Gaussians initiated from Structure-from-Motion (SfM)~\cite{schonberger2016structure} with learnable attributes such as color, shape, and opacity. These Gaussians, endowed with learnable shape and appearance parameters, can be splatted to 2D planes for rapid and differentiable rendering with rasterization~\cite{lassner2021pulsar}.
The advantages of rapid differentiable rendering with high photo-realistic fidelity have stimulated the fast and widespread adoption of 3DGS in the field.

Despite its high quality, 3DGS takes a lot of parameters because of its explicit expression.  Representing large scenes requires millions of neural Gaussian points, which demand a large amount of storage. Consequently, the substantial burden on storage and bandwidth hinders the practical applications of 3DGS and necessitates the development of compression methodologies. While an efficient compression method is required for 3DGS, the sparse and unorganized property of Gaussians makes it challenging. Various techniques are proposed to solve this problem. Some studies~\cite{fan2023lightgaussian,niedermayr2024compressed,navaneet2023compact3d,lee2024compact} focus on reducing memory consumption by clustering discrete, contiguous Gaussian attributes into codebooks. 

In contrast,~\cite{lu2024scaffold} introduces an MLP-based rendering process that achieves a tenfold reduction in size compared to vanilla 3DGS, while also enhancing fidelity and rendering quality.
Although Scaffold-GS~\cite{lu2024scaffold} presents a hierarchical and region-aware scene representation with a reliable anchor pruning and growing strategy that achieves a tenfold reduction in storage, it doesn't consider the relation of neural Gaussians (or anchors). Recent works~\cite{chen2024hac,wang2024contextgscompact3dgaussian} have revealed the background of inherent relations of unorganized anchors and compressed the neural Gaussians with structured hash grids.

\begin{figure}[t]
    \centering
    \includegraphics[width=0.9\linewidth]{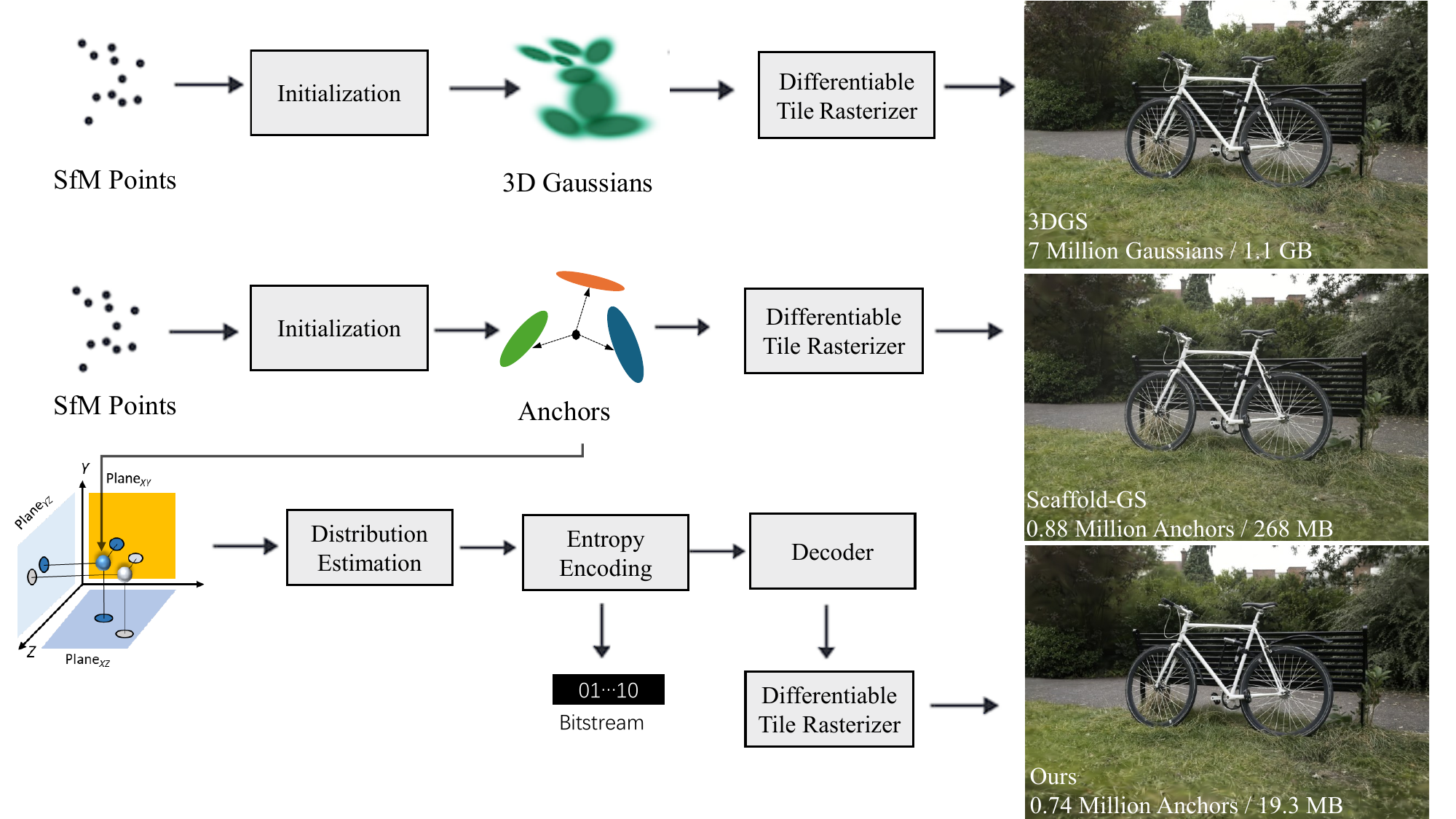}
     \vspace{-0.6em}
    \caption{\textbf{Comparison of two baselines and our model.} From top to bottom: 3DGS~\cite{kerbl20233d}, Scaffold-GS~\cite{lu2024scaffold} and TC-GS (ours). We conduct a simple experiment on the `bicycle' scene in the Mip-nerf360 dataset~\cite{barron2022mip}. The results highlight the superiority of TC-GS in achieving both reduced Gaussian quantity and significantly improved storage efficiency. This demonstrates the potential of our method to handle complex scenes with minimal resource requirements, making it a robust solution for scalable applications. }
    \label{fig:first}
    \vspace{-0.6em}
\end{figure}

While hash grids offer a discrete structure that reduces storage costs, their querying efficiency can be hindered by hash conflicts, which impact retrieval accuracy~\cite{xi2023neural,sun2024recent}. The tri-plane with continuous data structure can give smoother changes on high-resolution scenes, while the hash operation may involve discrete noise.
Inspired by~\cite{chan2022efficientgeometryaware3dgenerative}, we introduce a tri-plane-based compression framework, which is well-structured and easy-querying, where we jointly learn the tri-plane features and anchor attributes. Furthermore, we reveal an inherent relationship between the contiguous tri-plane and a cluster of anchors. Specifically, we address K-Nearest Neighbors (KNN)~\cite{peterson2009k} to anchors before the compression period. With entropy coding~\cite{witten1987arithmetic}, we can easily turn the sparse point cloud objects into bitstreams, dramatically reducing storage. Compared with baselines~\cite{lu2024scaffold,kerbl20233d} in~\cref{fig:first}, our model gains further optimization in storage efficiency.

Note that quantization in~\cite{chen2024hac,wang2024contextgscompact3dgaussian} can lead to blurs and floaters on object edges. To address that, we propose an adaptive learning loss based on wavelet~\cite{torrence1998practical} to make model attention on the high-frequency information in the scenes. Learnable masks are also employed to mask out invalid anchors considering scales and opacity, further increasing the compression efficiency. Our main contributions can be summarized as follows:
\begin{enumerate}
    \item With spatial correlations among neural Gaussians, we employ the Tri-plane structure for 3DGS compression, using projection and MLP to predict the inherent distribution of neural Gaussians (or anchors in Scaffold-GS).
    \item To enhance the capacity of the Tri-plane structure, we design a KNN Tri-plane decoder that decodes the tri-plane feature of the anchor along with the features of its K-nearest neighbor anchors. This KNN decoder performs better than decoding single anchor attributes in our experiments.
    \item To have good performance of high-frequency features of the scene, i.e., edges of objects, we propose an adaptive loss based on wavelet. We also employ learnable masks on anchors and neural Gaussians to prune the invisible ones during rendering progress.
    \item Our TC-GS has pioneered the Tri-plane-based compression method, achieving a $100\times$ storage reduction of vanilla 3DGS~\cite{kerbl20233d} and $14\times$ over Scaffold-GS~\cite{lu2024scaffold} while achieving comparable or even higher rendering quality.
\end{enumerate}

\section{Methodology}
\begin{figure*}[ht]
    \centering
    \vspace{-2.3em}
    \includegraphics[width=0.8\linewidth]{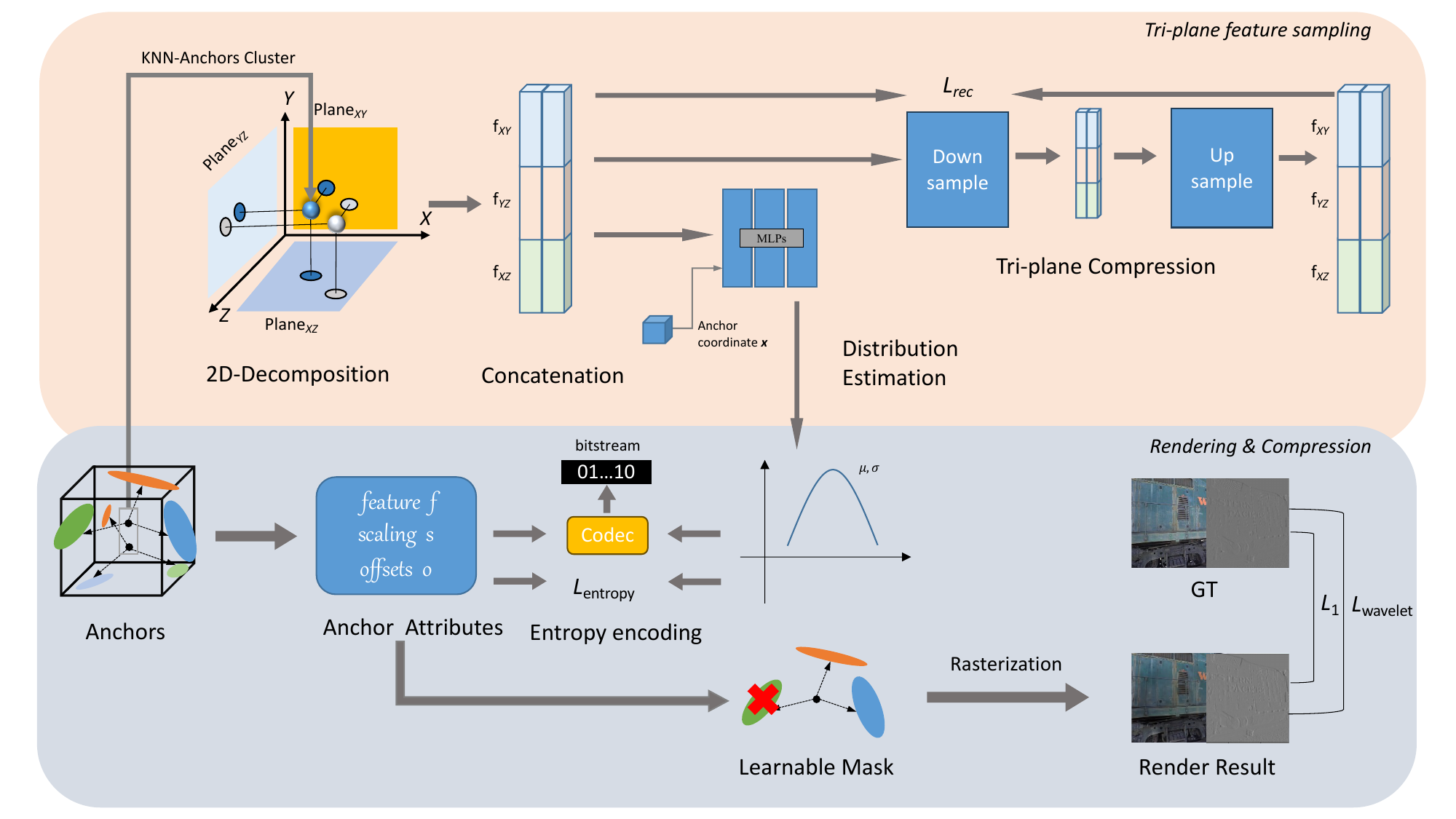}
    \vspace{-0.8em}
    \caption{\textbf{Overview of our model.} It follows Scaffold-GS~\cite{lu2024scaffold}, which introduces anchors to a compact representation of neural Gaussians. \textbf{Top left:} Our framework jointly learns contiguous Tri-plane while neural Gaussians rasterization and compressed with downsampling to reduce the storage cost of Tri-plane.  \textbf{Right:} Our context model exploits the output of Tri-plane as a context model to predict the distribution of anchor attributes. Then compressed with entropy encoding. \textbf{Bottom left:} To ensure high-frequency performance, e.g., edges of objects, we propose an adaptive wavelet constraint that leads the model to focus on low-frequency at the beginning and, as learning proceeds, on high-frequency features.
    }
    \label{fig:pipeline}
    \vspace{-0.6em}
\end{figure*}

The vanilla 3DGS~\cite{kerbl20233d} optimizes neural Guassians to reconstruct scenes but neglects the correlation between Gaussians. This can lead to abundant storage costs due to the redundancy of Gaussians. 
To this end, we propose our approach to solve the drawbacks of vanilla 3DGS.
In~\cref{fig:pipeline}, we illustrate our framework. On top of that, we sample features from our tri-plane to estimate the distribution of anchor attributes. On the bottom, we employ a rendering pipeline with entropy encoding to achieve compression.

In particular, our approach is based on the baseline Scaffold-GS~\cite{lu2024scaffold}.
To fully exploit the anchor-based framework, we introduce a Tri-plane representation leveraging the inherent space correlation among anchors to compress the unorganized Gaussians further. We first introduce the preliminaries of both 3DGS and Scaffold-GS in \cref{preliminaries}. Based on that, we present the detailed technical components of our framework. In \cref{compress}, we present our Tri-plane-based context model, which is utilized to compress anchor features. \cref{mask}  describes a masking method that reduces excessive Gaussian storage costs. In \cref{tri-compression}, we propose a technique to further compress Tri-plane storage. Finally, in \cref{waveletconstraint}, we introduce an adaptive wavelet constraint aimed at mitigating high-frequency information loss during compression.

\subsection{Preliminaries} \label{preliminaries}
\textbf{3DGS}~\cite{kerbl20233d} utilizes a collection of anisotropic 3D neural Gaussians to depict the scene so that the scene can be efficiently rendered using a tile-based rasterization technique. Beginning from a set of Structure-from-Motion (SfM) points, each Gaussian point is represented as follows:
\begin{equation}
G(x) = e^{-\frac{1}{2}(x-\mu)^\top \Sigma^{-1} (x-\mu)} ,
\end{equation}
where $x$ is an arbitrary position within the 3D scene and $\Sigma$ denotes the covariance matrix of the 3D Gaussian. $\Sigma$ is formulated using a scaling matrix $S$ and rotation matrix $R$ to maintain its positive semi-definite:
\begin{equation}
\Sigma = R S S^\top R^\top .
\end{equation}

All the attributes, i.e., $[\mu, R, S, \alpha, c]$, in neural Gaussian points are learnable and optimized by the reconstruction loss.

\textbf{Scaffold-GS}~\cite{lu2024scaffold} proposes to introduce an \textit{anchor}-based pipeline with impressive storage deduction without sacrificing high-fidelity. 
Instead of directly storing attributes, the attributes, i.e., color $c$, opacity $\alpha$, location $x$, rotation $r$, and scaling $s$, are predicted from the attributes of attached anchors through MLPs. Each anchor point has a context feature $f \in \mathbb{R}^{32}$, a location $x \in \mathbb{R}^{3}$, a scaling factor $l \in \mathbb{R}^{3}$ and \textit{k} learnable offset $O \in \mathbb{R}^{k\times3}$.

While Scaffold-GS has demonstrated effectiveness via this anchor-centered design, we contend there is still significant redundancy among inherent consistencies and correlations of anchors that we can further exploit for a more compact 3DGS representation.

\subsection{Tri-plane Context Model for Compression} \label{compress}
The main idea of our work is to exploit the continuous and structured Tri-plane to gain the correlation among sparse neural Gaussians and then utilize it to fully compress Gaussians. Furthermore, with 2D well-structured properties, Tri-plane can be easily compressed and stored with little storage consumption. To starters, inspired by~\cite{chan2022efficientgeometryaware3dgenerative}, we directly substitute Tri-plane features for the anchor feature. However, the results show that direct substitution leads to unstable training and hard sampling for the growing stage of anchors. There is also a degradation of fidelity because the sampling on the Tri-plane results in the blending of anchor features.

In this circumstance, we adopt the pipeline based on Scaffold-GS~\cite{lu2024scaffold} and employ Tri-plane as a prior of anchor features. In previous works~\cite{chen2024hac,wang2024contextgscompact3dgaussian}, they find that all three components of anchors \{$f, l, o$\} exhibit statistical Gaussian distributions.
The goal of our approach is to leverage the Tri-plane feature to predict the distribution of anchor components, and then use the entropy model to compress the attributes.

To get features from the Tri-plane, we first project anchor coordinates from 3D to 2D by the axis of the Tri-plane and employ bi-linear interpolation in three planes, i.e., \textit{xy, yz, xz} planes.
What needs to be emphasized is that we cannot define the coordinate boundary for Tri-planes as the same as the Gaussian radius. Thus, we need a scaling method to schedule the maximum of 2D coordinates. We utilize the contract function in~\cite{barron2022mip}, which guarantees the Gaussian coordinates lie within the boundary of the Tri-plane. In~\cref{eq:contract}, we first normalize coordinates and contract coordinates within $[0,1]$. This method of sampling from Tri-planes performs better than merely contracting by maximal coordinates. Specifically, we use the minimum between scene radius from Sfm~\cite{schonberger2016structure} and the Bbox in iteration $10,000$.
\begin{equation}
\text{contract}(\mathbf{x}) = 
\begin{cases} 
\mathbf{x},  & \|\mathbf{x}\| \leq 1 ,\\
\left( 2 - \frac{1}{\|\mathbf{x}\|} \right) \left( \frac{\mathbf{x}}{\|\mathbf{x}\|} \right),  & \|\mathbf{x}\| > 1 .
\end{cases}
\label{eq:contract}
\end{equation}

Then we apply view-dependent MLP to predict the distribution of Gaussian attributes. While it performs well in this way, we explore that there are specific connections between nearby anchors in space. A KNN-clustering step is employed to leverage this, and we predict distribution from the anchor and its K-Nearest Neighbour. Moreover, we add quantization to facilitate the entropy encoding for the attributes, which should be a finite set. 

Specifically, we follow the technique introduced by~\cite{chen2024hac} to calculate the bit consumption of prediction of the Gaussians' attributes. Utilizing a view-based MLP to predict the $\mu$ and $\sigma$ from the clustered and anchor features guides the model to fit quantized attributes.
Consequently, we define an entropy loss as the summation of bit consumption overall $\hat{f}_{i}$s:
\begin{equation}
\mathcal{L}_{\text{entropy}} = \sum_{f \in \{f^a, l, o\}} \sum_{i=1}^{N} \sum_{j=1}^{D} \left( -\log_2 p(\hat{f}_{i,j}) \right) ,
\end{equation}
where N is the number of anchors and $\hat{f}_{i,j}$ is $j$-th dimension value of $\hat{f}_{i}$.
With the effort of entropy loss, we can obtain a more accurate prediction of distributions.

\subsection{Anchor Masking} \label{mask}
With an abundant number of neural Gaussians in the scene, some of them do not perform during the rasterization stage, suggesting the occurrence of substantial unnecessary Gaussians. This redundancy not only increases computational overhead but also leads to inefficient memory utilization in the neural representation. 
To address this issue, we deploy adaptive anchor masks based on~\cite{lee2024compact}. To eliminate anchors and neural Gaussians who don't contribute in rasterization, we prune invalid offset by utilizing straight-through~\cite{bengio2013estimatingpropagatinggradientsstochastic} estimated binary masks. 
In this way, we can effectively delete invalid offsets and save storage, resulting in a more compact and efficient neural representation without sacrificing the visual quality of the rendered scenes.

\subsection{Compression of Tri-plane} \label{tri-compression}
Unlike voxels and hash grids, Tri-plane has a contiguous 2D structure in space instead of a sparse one. Thus, we can employ some intuitive compression for 2D data structure.
Inspired by U-net~\cite{ronneberger2015unetconvolutionalnetworksbiomedical}, we design a convolutional network that first downsamples to compressed latent and upsamples back to the input. As the training progresses, we use the entire network after the Tri-plane output to learn from the Tri-plane feature, as~\cref{fig:pipeline} shows. After that, we only save the compressed latent and upsampling decoder for inference to reduce the storage cost. In this case, the storage of Tri-plane can be reduced by 5$\times$ than directly saved. Specifically, we supervise the compression of the Tri-plane with $L_1$ loss between the original and reconstructed feature, as shown in~\cref{eq:tri_compress}
\begin{equation}
\mathcal{L}_{tri\_rec} =  L_1(f_{original}, f_{reconsturcted})  .
\label{eq:tri_compress}
\end{equation}

\subsection{Adaptive Wavelet Constraint} \label{waveletconstraint}
While quantization contributes significantly to compression overall, it also makes blurs and artifacts from the edges of objects. To address this, we employ an adaptive loss based on wavelet transform~\cite{torrence1998practical}, which allows the model to focus on broad aspects first. As iterations progress, concentrate on detailed features. Specifically, we utilize a discrete wavelet transform with two-level decomposition. We activate $L_1$ loss based on the frequency map from the rendered image and ground truth. Then calculate with~\cref{eq:wavelet} to regularize the model.
\begin{equation}
\mathcal{L}_{wavelet} = \lambda_1 \cdot L_1(\text{YL}(x_1), \text{YL}(x_2)) + \lambda_2 \cdot L_1(\text{YH}(x_1), \text{YH}(x_2)) , 
\label{eq:wavelet}
\end{equation}
which consists of two terms: \textbf{YL} for the low-frequency components and \textbf{YH} for the high-frequency components of the input images $x_1$ and $x_2$. The weights of these components, $\lambda_1$ and $\lambda_2$ dynamically adjust based on the training step.

During training, we incorporate both the rendering fidelity loss and the entropy loss to ensure the model improves rendering quality while regularizing frequency features with wavelet transform. Our overall loss $\mathcal{L}_{\text{all}}$ can be formulated as
\begin{equation}
\mathcal{L}_{\text{all}} = \mathcal{L}_{\text{Scaffold}} + \lambda_e \cdot \frac{1}{\epsilon}\mathcal{L}_{\text{entropy}}  + \lambda_m \mathcal{L}_m + \lambda_w \mathcal{L}_{wavelet} + \lambda_{tc} \mathcal{L}_{tri\_rec}, 
\label{eq:loss}
\end{equation}
where  $\mathcal{L}_{\text{Scaffold}}$ stands for loss defined in~\cite{lu2024scaffold} and $\mathcal{L}_m$ represents mask loss in~\cite{lee2024compact}. The $\mathcal{L}_{\text{entropy}}$ in~\cref{eq:loss} is wavelet loss defined in~\cref{eq:wavelet} and $\epsilon$ is a scaling factor to regularize Tri-plane learning.
$\lambda_m$, $\lambda_e$, $\lambda_w$ and $\lambda_{tc}$  are trade-off hyperparameters used
to balance the loss components.

\begin{table*}[ht] \scriptsize
    \centering
    \setlength\tabcolsep{1.4pt}
    \caption{Quantitative results. 3DGS~\cite{kerbl20233d} and Scaffold-GS~\cite{lu2024scaffold} are two baselines. Approaches in the middle chunk are designed for 3DGS compression.
    The best and 2nd best results are highlighted in \colorbox{red!25}{red} and \colorbox{yellow!25}{yellow} cells respectively. The size is measured in MB.
    }
     \vspace{-0.8em}
    \resizebox{0.6\textwidth}{!}{
    \begin{tabular}{ll|cccc|cccc|cccc}
        \toprule
        \multicolumn{2}{l|}{\textbf{Datasets}}          & \multicolumn{4}{c|}{\textbf{Synthetic-NeRF}}  & \multicolumn{4}{c}{\textbf{Tank\&Temple}}
        & \multicolumn{4}{c}{\textbf{DeepBlending}}\\
        \multicolumn{2}{l|}{\textbf{Methods}} & PSNR$\uparrow$    & SSIM$\uparrow$   & LPIPS$\downarrow$ & SIZE$\downarrow$   & PSNR$\uparrow$    & SSIM$\uparrow$   & LPIPS$\downarrow$ & SIZE$\downarrow$   & PSNR$\uparrow$   & SSIM$\uparrow$   & LPIPS$\downarrow$ & SIZE$\downarrow$   \\
        \bottomrule
        \multicolumn{2}{l|}{\textbf{3DGS}}& \colorbox{red!25}{33.80}&\colorbox{red!25} {0.970}&\colorbox{red!25} {0.031}&68.46& 23.69&0.844&0.178&431.0&29.42&0.899&0.247&663.9   \\
        \multicolumn{2}{l|}{\textbf{Scaffold-GS}}&33.41&0.966&0.035&19.36&23.96& \colorbox{red!25}{0.853}&  \colorbox{yellow!25}{0.177}&86.50&  \colorbox{yellow!25}{30.21}& \colorbox{yellow!25}{0.906}&0.254&66.00    \\  \hline
        \multicolumn{2}{l|}{\textbf{Lee}}&33.33&  \colorbox{yellow!25}{0.968}&0.034&5.54&23.32&0.831&0.201&39.43&29.79&0.901&0.258&43.21    \\  
        \multicolumn{2}{l|}{\textbf{Compressed3D}}&32.94&0.967&  \colorbox{yellow!25}{0.033}&3.68&23.32&0.832&0.194&17.28&29.38&0.898&0.253&25.30    \\  
        \multicolumn{2}{l|}{\textbf{EAGLES}}&32.54&0.965&0.039&5.74&23.41&0.840&0.200&34.00&29.91& \colorbox{red!25}{0.910}&0.250&62.00    \\
        \multicolumn{2}{l|}{\textbf{LightGaussian}}&32.73&0.965&0.037&7.84&22.83&0.822&0.242&22.43&27.01&0.872&0.308&33.94    \\  
        \multicolumn{2}{l|}{\textbf{Morgen.}}&31.05&0.955&0.047&2.20&22.78&0.817&0.211&13.05&28.92&0.891&0.276&8.40    \\
        \multicolumn{2}{l|}{\textbf{Navaneet}}&33.09& {0.967}&0.036&4.42&23.47&0.840&0.188&27.97&29.75&0.903&  \colorbox{yellow!25}{0.247}&42.77    \\  
        \multicolumn{2}{l|}{\textbf{HAC}}&  \colorbox{yellow!25}{33.71}&  \colorbox{yellow!25}{0.968}&0.034&  \colorbox{yellow!25}{1.86}& \colorbox{red!25}{24.40}& \colorbox{red!25}{0.853}& {0.17 7}& {11.24}&\colorbox{red!25}{30.34}&  \colorbox{yellow!25}{0.906}&0.258&  \colorbox{yellow!25}{6.35}   \\  \hline
        \multicolumn{2}{l|}{\textbf{TC-GS(Ours)}}&31.28&0.957&0.034& \colorbox{red!25}{1.33}& {23.96}&0.843& \colorbox{red!25}{0.115}& \colorbox{red!25}{7.66}&30.07&0.902& \colorbox{red!25}{0.121}& \colorbox{red!25}{3.77}   \\ 
        \hline
        \toprule
    \end{tabular}
    }
    \label{tab:main_quantitative}
    \vspace{-1.5em}
\end{table*}

\section{Experiments}

In this section, we first present our framework’s implementation details and then conduct evaluation experiments to compare with existing 3DGS compression approaches. Additionally, we include ablation studies to demonstrate the effectiveness of each technical component of our method.

\subsection{Implementation Details}
To utilize a stable compression process, we start training with vanilla Scaffold-GS~\cite{lu2024scaffold}. After anchors stop growing, we add quantization and the Tri-plane context model to estimate the distribution of anchor attributes. We employ upsample and downsample modules to compress the tri-plane. 
Subsequently, we leverage the entropy encoding/decoding process with codec.

Specifically, we implement our model based on Scaffold-GS~\cite{lu2024scaffold} implementation in PyTorch framework.
We utilize the Tri-plane with random noise. We set the learning rate of the Tri-plane from $5e-3$ to $1e-5$ as iterations going. For the upsample and downsample modules, we define a single three 2D-convolution layers MLP with ReLU activation. The learning rate of the adaptive anchor mask is set from $1e-2$ to $1e-4$. And the K of KNN clustering is set to 4. Other hyper-parameter settings follow Scaffold-GS~\cite{lu2024scaffold}.

\subsection{Experiment Evaluation}
\noindent\textbf{Baselines}. We compare our model with existing 3DGS compression approaches. Notably,~\cite{fan2023lightgaussian, lee2024compact, navaneet2023compact3d, niedermayr2024compressed} mainly adopt codebook-based or parameter pruning strategies, while Scaffold-GS~\cite{lu2024scaffold} explores Gaussian relations for compact representation. Additionally, EAGLES~\cite{girish2024eaglesefficientaccelerated3d} and Morgenstern~\cite{morgenstern2024compact3dscenerepresentation} employ non-contextual entropy constraints and dimension collapse techniques, respectively.

\vspace{0.3em}
\noindent\textbf{Datasets and Evaluation Metrics}. We evaluate our approach on diverse datasets, including the small-scale Synthetic-NeRF~\cite{mildenhall2020nerfrepresentingscenesneural} and the large-scale real-scene datasets DeepBlending~\cite{deepblending} and Tanks \& Temples~\cite{tankstemple}, demonstrating its effectiveness across scenarios. We evaluate image quality using PSNR, SSIM, and LPIPS. 

\begin{figure}[t]
    \centering
    \includegraphics[width=0.90\linewidth]{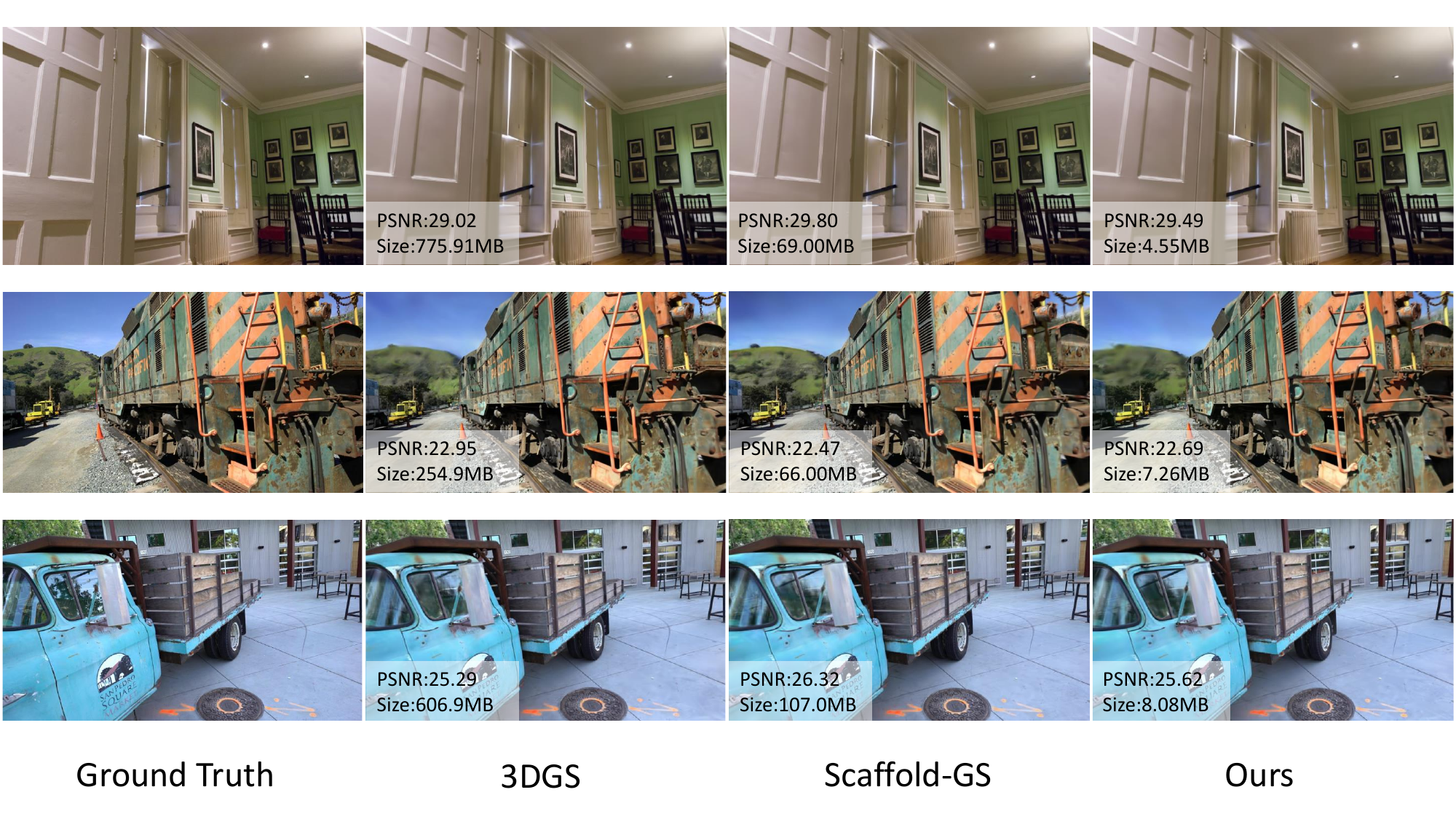}
    \vspace{-1.5em}
    \caption{\textbf{Qualitative Results} on ``train'' and ``truck'' from Tanks and temples~\cite{tankstemple} and ``drjohnson'' from DeepBlending~\cite{deepblending}. The PSNR and storage costs are given on the lower left.}
    \label{fig:main_qualitative}
    \vspace{-0.6em}
\end{figure}

\vspace{0.3em}
\noindent\textbf{Quantitative Results}. Quantitative results are shown in ~\cref{tab:main_quantitative}, and the qualitative outputs are presented in~\cref{fig:main_qualitative}. Our model has achieved a significant storage reduction of over $100\times$ compared to the vanilla 3DGS~\cite{kerbl20233d} with improved fidelity. The size reduction also exceeds $17\times$ over the base model Scaffold-GS~\cite{lu2024scaffold}  with equal fidelity. Notably, our lpips metric surpasses other models, primarily thanks to the adaptive wavelet regularization. Although other compression approaches can reduce the model size by primarily using pruning and codebooks, they still exhibit significant spatial redundancy. Among the bounded and unbounded scene datasets~\cite{tankstemple,mildenhall2020nerfrepresentingscenesneural,deepblending}, we all achieve the highest compression rate by effectively leveraging the well-structured Tri-plane representation. This approach not only optimizes storage efficiency but also ensures high fidelity in visual quality, as evidenced by a consistently low LPIPS score. Such results highlight the effectiveness of our method in balancing compression and perceptual quality, making it a robust solution for large-scale scene representation.

\begin{figure}[t]
    \centering
    \includegraphics[width=0.90\linewidth]{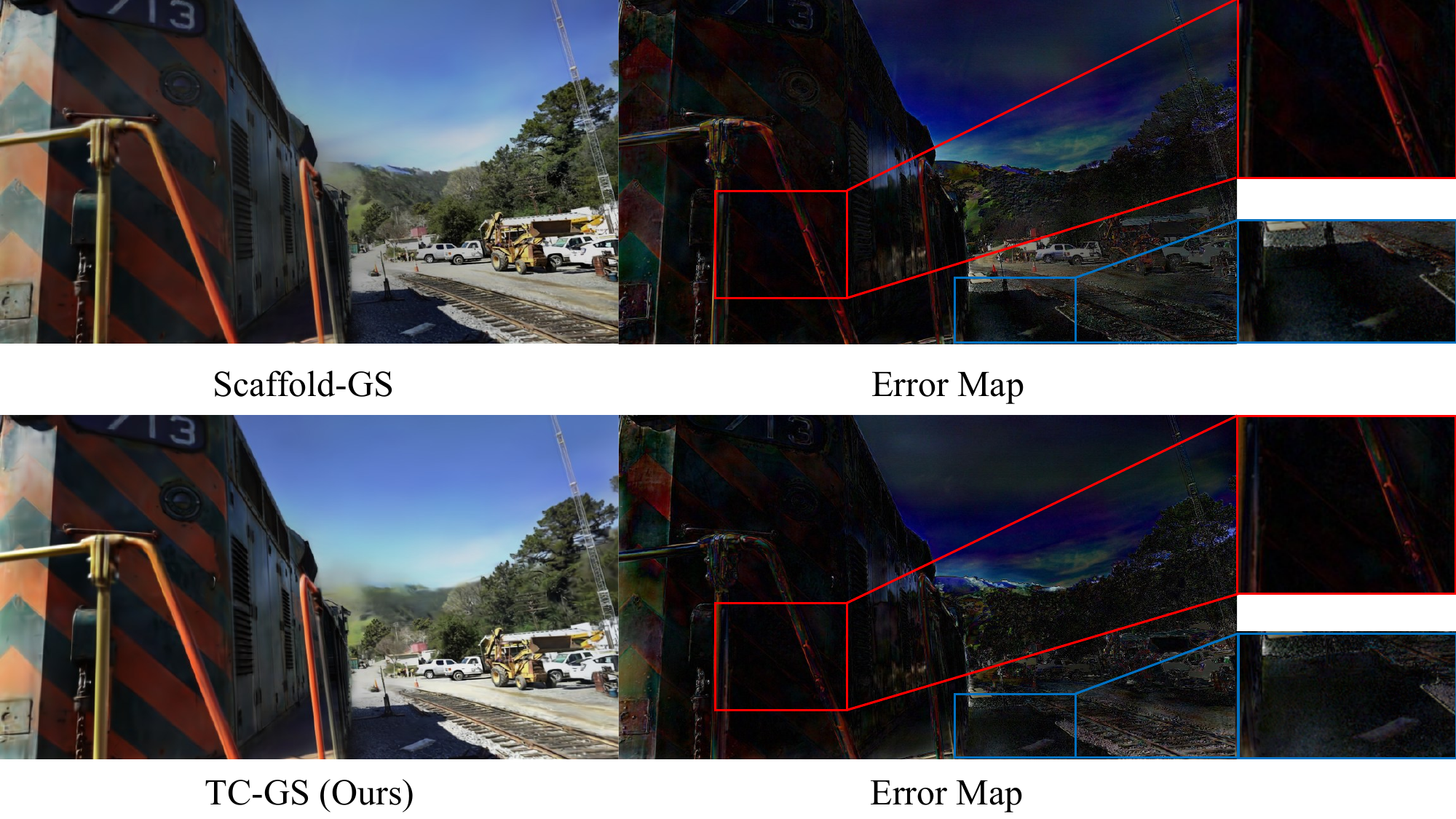}
    \vspace{-1.1em}\caption{\textbf{Qualitative comparisons} on the `train' scene from Tanks and Temples~\cite{tankstemple}. The left column presents rendered images from novel views, while the middle column shows the corresponding error maps, computed as the absolute pixel-wise differences between the rendered outputs and the ground truth. In these error maps, darker areas indicate higher accuracy, i.e., smaller deviations indicate better rendering quality. Our approach demonstrates superior performance, not only in faithfully reproducing the main item but also in capturing intricate details and substructures, such as the rail and shadows, surpassing the baseline method~\cite{lu2024scaffold}. Note that we brighten up the error maps to enhance visibility.}
    
    \label{fig:qualitative_result_w/err}
    \vspace{-0.8em}
\end{figure}

\vspace{0.3em}
\noindent\textbf{Qualitative Results}. As shown in \cref{fig:main_qualitative}, our model keeps high fidelity with an incredible storage reduction. By leveraging the Tri-plane as prior information, our model demonstrates exceptional performance in generating high-quality rendered images. This structured representation serves as a powerful foundation, enabling the model to capture intricate scene details and maintain visual consistency across various viewpoints. Furthermore, we conduct a comparison between our model and Scaffold-GS~\cite{lu2024scaffold} on the `train' scene from Tanks and Temples~\cite{tankstemple}, as presented in~\cref{fig:qualitative_result_w/err}. As illustrated by the error maps, our model demonstrates superior accuracy in both color reproduction and geometric shape, reflecting its ability to achieve higher fidelity in rendering. With the help of adaptive wavelet constraint, our model results in better performance in edges and object shapes. Additionally, the wavelet assists in scene learning without floaters and blur. The qualitative results show that our approach has achieved effective performance between the trade-off of compression ratio and rendering fidelity. 

\subsection{Ablation Study}    
In this subsection, we conduct ablation studies to demonstrate the effectiveness of each technical component in our framework. We conduct our experiments on unbounded Tanks \& Temple dataset~\cite{tankstemple} to support convincing results.
We assess the effectiveness of each technical component by disabling either of the following: 1) correlation information stored in Tri-plane; 2) compression of Tri-plane; and 3) adaptive wavelet constraint. The results are shown in \cref{tab:ablation}.

\vspace{0.3em}
\noindent\textbf{Tri-plane Context Model.} We employ the \textbf{w/o tri-plane} scenario by directly substituting the anchor feature from a Tri-plane instead of learning the correlation among anchors. Although the Tri-plane with high-dimension features can achieve the same effectiveness as anchors, the results show that the storage cost times than using the context model, which demonstrates that the Tri-plane context model within our framework excels at extracting valuable information from the correlation among anchors, thereby achieving an impressive compression ratio. 

\noindent\textbf{Compression of Tri-plane.} As shown in~\cref{tab:ablation}, the Tri-plane structure with a high feature dimension leads to significant storage costs. To address this, we leverage the process of upsampling and downsampling to maintain a compressed latent representation, effectively reducing storage requirements while retaining critical information. Furthermore, by integrating compression learning directly into the Tri-plane compression module, we enhance the efficiency of the compression process. This approach not only mitigates the high storage costs associated with the Tri-plane's high feature dimensions but also achieves superior compression results without compromising rendering quality, enabling high-fidelity scene representation with minimal storage overhead.

\vspace{0.3em}
\noindent\textbf{Adaptive Wavelet Constraint.} The proposed wavelet constraint provides the model with an additional mechanism to enhance its focus on details in rendered scenes. By incorporating multi-scale frequency information, this constraint enables the model to effectively capture and preserve intricate textures and structural nuances, resulting in more realistic and visually appealing outputs.  When the adaptive wavelet constraint is omitted from our training process, we observe a noticeable drop in rendering quality. 
This outcome highlights the effectiveness and essential role of our component in maintaining high-quality rendering.

\vspace{0.3em}
\noindent\textbf{Anchor Mask.} The anchor mask functions as an effective filter between the training and encoding stages, ensuring that only relevant anchors are retained for the rendering process. One of the most straightforward methods to reduce time and storage costs is by minimizing the number of anchors. Without this filtering mechanism provided by the anchor mask, the number of anchors grows indiscriminately, including many that do not contribute meaningfully to the rendering process. As a result, the overall scene size expands unnecessarily, leading to inefficiencies in both storage and computational requirements.

\begin{table}[t]
    \centering
    \vspace{-0.8em}
    \setlength\tabcolsep{1.6pt}
    \caption{Ablations of different components in our framework.}
     \vspace{-0.8em}
    \resizebox{0.7\linewidth}{!}{%

    \begin{tabular}{l|cccc}
    \hline
    \toprule
          & PSNR$\uparrow$ & SSIM$\uparrow$ & LPIPS$\downarrow$ & SIZE$\downarrow$ \\ \hline
        full & 22.31 & 0.81 & 0.15 & 7.11 \\ \hline
        w/o tri-plane & 22.30 & 0.80 & 0.15 & 47.3645 \\ \hline
        w/o tri-compression & 22.31 & 0.81 & 0.16 & 8.01 \\ \hline
        w/o wavelet & 22.13 & 0.81 & 0.17 & 7.26 \\ \hline
        w/o anchor mask & 22.28 & 0.81 & 0.15 & 7.40 \\ \hline
        \toprule
    \end{tabular}
    }
    \label{tab:ablation}
    \vspace{-2.3em}
\end{table}

\vspace{-0.3em}
\subsection{Limitation}
The use of additional models in our framework can result in increased training time, approximately $1.88\times$ longer than Scaffold-GS.
For the unified scene `train' in~\cite{tankstemple}, the training times are 38 $mins$ for 3DGS~\cite{kerbl20233d}, 25 $mins$ for Scaffold-GS~\cite{lu2024scaffold} and 45 $mins$ for our framework. Specifically, the K-NN clustering spends most of the training time and leads to a drop in rendering FPS.
The encoding/decoding process takes approximately 29 seconds on the scene 'train' in~\cite{tankstemple}. The dominant time consumption occurs during Codec execution of entropy encoding on the CPU, as we follow the implementation of~\cite{chen2024hac}.

\vspace{-0.2em}
\section{Conclusion}
In this paper, we have presented a pioneering study that explores the integration of a Tri-plane-based context model within 3D Gaussian splatting (3DGS) models. By leveraging the well-structured and continuous nature of the Tri-plane, we are able to efficiently determine the distribution of Gaussian (or anchor) attributes, enabling the application of entropy encoding techniques to significantly reduce storage costs. However, while the proposed method achieves notable compression efficiency, the rendering quality is affected by the inevitable compression and quantization. This highlights the need for further advancements to refine the balance between high-quality rendering and optimal compression, paving the way for more robust and scalable solutions in future work.

\bibliographystyle{IEEEbib}
\bibliography{references}

\begin{thebibliography}{10}

\bibitem{mildenhall2020nerfrepresentingscenesneural}
Ben Mildenhall, Pratul~P. Srinivasan, Matthew Tancik, Jonathan~T. Barron, Ravi Ramamoorthi, and Ren Ng,
\newblock ``Nerf: Representing scenes as neural radiance fields for view synthesis,''
\newblock in {\em ECCV}, 2020.

\bibitem{muller2022instant}
Thomas M{\"u}ller, Alex Evans, Christoph Schied, and Alexander Keller,
\newblock ``Instant neural graphics primitives with a multiresolution hash encoding,''
\newblock {\em ACM TOG}, vol. 41, no. 4, pp. 1--15, 2022.

\bibitem{barron2021mip}
Jonathan~T Barron, Ben Mildenhall, Matthew Tancik, Peter Hedman, Ricardo Martin-Brualla, and Pratul~P Srinivasan,
\newblock ``Mip-nerf: A multiscale representation for anti-aliasing neural radiance fields,''
\newblock in {\em ICCV}, 2021, pp. 5855--5864.

\bibitem{barron2022mip}
Jonathan~T Barron, Ben Mildenhall, Dor Verbin, Pratul~P Srinivasan, and Peter Hedman,
\newblock ``Mip-nerf 360: Unbounded anti-aliased neural radiance fields,''
\newblock in {\em CVPR}, 2022, pp. 5470--5479.

\bibitem{hu2023tri}
Wenbo Hu, Yuling Wang, Lin Ma, Bangbang Yang, Lin Gao, Xiao Liu, and Yuewen Ma,
\newblock ``Tri-miprf: Tri-mip representation for efficient anti-aliasing neural radiance fields,''
\newblock in {\em ICCV}, 2023, pp. 19774--19783.

\bibitem{kerbl20233d}
Bernhard Kerbl, Georgios Kopanas, Thomas Leimk{\"u}hler, and George Drettakis,
\newblock ``3d gaussian splatting for real-time radiance field rendering.,''
\newblock {\em ACM Trans. Graph.}, vol. 42, no. 4, pp. 139--1, 2023.

\bibitem{schonberger2016structure}
Johannes~L Schonberger and Jan-Michael Frahm,
\newblock ``Structure-from-motion revisited,''
\newblock in {\em CVPR}, 2016, pp. 4104--4113.

\bibitem{lassner2021pulsar}
Christoph Lassner and Michael Zollhofer,
\newblock ``Pulsar: Efficient sphere-based neural rendering,''
\newblock in {\em CVPR}, 2021, pp. 1440--1449.

\bibitem{fan2023lightgaussian}
Zhiwen Fan, Kevin Wang, Kairun Wen, Zehao Zhu, Dejia Xu, and Zhangyang Wang,
\newblock ``Lightgaussian: Unbounded 3d gaussian compression with 15x reduction and 200+ fps,''
\newblock {\em arXiv preprint arXiv:2311.17245}, 2023.

\bibitem{niedermayr2024compressed}
Simon Niedermayr, Josef Stumpfegger, and R{\"u}diger Westermann,
\newblock ``Compressed 3d gaussian splatting for accelerated novel view synthesis,''
\newblock in {\em CVPR}, 2024, pp. 10349--10358.

\bibitem{navaneet2023compact3d}
KL~Navaneet, Kossar~Pourahmadi Meibodi, Soroush~Abbasi Koohpayegani, and Hamed Pirsiavash,
\newblock ``Compact3d: Compressing gaussian splat radiance field models with vector quantization,''
\newblock {\em arXiv preprint arXiv:2311.18159}, 2023.

\bibitem{lee2024compact}
Joo~Chan Lee, Daniel Rho, Xiangyu Sun, Jong~Hwan Ko, and Eunbyung Park,
\newblock ``Compact 3d gaussian representation for radiance field,''
\newblock in {\em CVPR}, 2024, pp. 21719--21728.

\bibitem{lu2024scaffold}
Tao Lu, Mulin Yu, Linning Xu, Yuanbo Xiangli, Limin Wang, Dahua Lin, and Bo~Dai,
\newblock ``Scaffold-gs: Structured 3d gaussians for view-adaptive rendering,''
\newblock in {\em CVPR}, 2024, pp. 20654--20664.

\bibitem{chen2024hac}
Yihang Chen, Qianyi Wu, Jianfei Cai, Mehrtash Harandi, and Weiyao Lin,
\newblock ``Hac: Hash-grid assisted context for 3d gaussian splatting compression,''
\newblock {\em arXiv preprint arXiv:2403.14530}, 2024.

\bibitem{wang2024contextgscompact3dgaussian}
Yufei Wang, Zhihao Li, Lanqing Guo, Wenhan Yang, Alex~C Kot, and Bihan Wen,
\newblock ``Contextgs: Compact 3d gaussian splatting with anchor level context model,''
\newblock {\em arXiv preprint arXiv:2405.20721}, 2024.

\bibitem{xi2023neural}
Yang Xi, Wanna Luan, and Jun Tao,
\newblock ``Neural monte carlo rendering of finite-time lyapunov exponent fields,''
\newblock {\em Visual Intelligence}, vol. 1, no. 1, pp. 10, 2023.

\bibitem{sun2024recent}
Jia-Mu Sun, Tong Wu, and Lin Gao,
\newblock ``Recent advances in implicit representation-based 3d shape generation,''
\newblock {\em Visual Intelligence}, vol. 2, no. 1, pp. 9, 2024.

\bibitem{chan2022efficientgeometryaware3dgenerative}
Eric~R Chan, Connor~Z Lin, Matthew~A Chan, Koki Nagano, Boxiao Pan, Shalini De~Mello, Orazio Gallo, Leonidas~J Guibas, Jonathan Tremblay, Sameh Khamis, et~al.,
\newblock ``Efficient geometry-aware 3d generative adversarial networks,''
\newblock in {\em CVPR}, 2022, pp. 16123--16133.

\bibitem{peterson2009k}
Leif~E Peterson,
\newblock ``K-nearest neighbor,''
\newblock {\em Scholarpedia}, vol. 4, no. 2, pp. 1883, 2009.

\bibitem{witten1987arithmetic}
Ian~H Witten, Radford~M Neal, and John~G Cleary,
\newblock ``Arithmetic coding for data compression,''
\newblock {\em Communications of the ACM}, vol. 30, no. 6, pp. 520--540, 1987.

\bibitem{torrence1998practical}
Christopher Torrence and Gilbert~P Compo,
\newblock ``A practical guide to wavelet analysis,''
\newblock {\em Bulletin of the American Meteorological society}, vol. 79, no. 1, pp. 61--78, 1998.

\bibitem{bengio2013estimatingpropagatinggradientsstochastic}
Yoshua Bengio, Nicholas L{\'e}onard, and Aaron Courville,
\newblock ``Estimating or propagating gradients through stochastic neurons for conditional computation,''
\newblock {\em arXiv preprint arXiv:1308.3432}, 2013.

\bibitem{ronneberger2015unetconvolutionalnetworksbiomedical}
Olaf Ronneberger, Philipp Fischer, and Thomas Brox,
\newblock ``U-net: Convolutional networks for biomedical image segmentation,''
\newblock in {\em MICCAI}. Springer, 2015, pp. 234--241.

\bibitem{girish2024eaglesefficientaccelerated3d}
Sharath Girish, Kamal Gupta, and Abhinav Shrivastava,
\newblock ``Eagles: Efficient accelerated 3d gaussians with lightweight encodings,''
\newblock in {\em ECCV}. Springer, 2025, pp. 54--71.

\bibitem{morgenstern2024compact3dscenerepresentation}
Wieland Morgenstern, Florian Barthel, Anna Hilsmann, and Peter Eisert,
\newblock ``Compact 3d scene representation via self-organizing gaussian grids,''
\newblock in {\em ECCV}. Springer, 2025, pp. 18--34.

\bibitem{deepblending}
Peter Hedman, Julien Philip, True Price, Jan-Michael Frahm, George Drettakis, and Gabriel Brostow,
\newblock ``Deep blending for free-viewpoint image-based rendering,''
\newblock {\em ACM Trans. Graph.}, vol. 37, no. 6, Dec. 2018.

\bibitem{tankstemple}
Arno Knapitsch, Jaesik Park, Qian-Yi Zhou, and Vladlen Koltun,
\newblock ``Tanks and temples: benchmarking large-scale scene reconstruction,''
\newblock {\em ACM Trans. Graph.}, vol. 36, no. 4, July 2017.

\end{thebibliography}

\end{document}